\documentclass[11pt,a4paper]{article}

\usepackage[T1]{fontenc}
\usepackage[british]{babel}
\usepackage[a4paper,margin=25mm,includefoot]{geometry}
\usepackage{amsmath,amssymb}
\usepackage{newtxtext,newtxmath}
\usepackage{microtype}
\usepackage{booktabs,tabularx,siunitx}
\usepackage[numbers,sort&compress]{natbib}
\usepackage{xcolor}
\definecolor{linkblue}{HTML}{193A62}
\usepackage[unicode,colorlinks,allcolors=linkblue]{hyperref}
\title{Effect of class proportion shifts on classifier curvature}
\author{Maksim Antonovich Bolshim\thanks{Correspondence: \href{mailto:mabolshim@itmo.ru}{mabolshim@itmo.ru}.}
  \quad and \quad Alexander Vladimirovich Kugaevskikh\\[5pt]
  \small ITMO University, Saint Petersburg, Russia}
\date{}
\hypersetup{
  pdftitle={Effect of class proportion shifts on classifier curvature},
  pdfauthor={Maksim Antonovich Bolshim and Alexander Vladimirovich Kugaevskikh},
  pdfsubject={An empirical study of the loss Hessian trace during short classifier fine-tuning},
  pdfkeywords={class proportions, Hessian trace, fine-tuning, classifier, gradient forecast}
}

\begin{document}
\maketitle
\begin{abstract}
Changing the class mix can alter mean loss without changing the classifier. To distinguish this reweighting from changes in the model itself, we studied ten-class CIFAR-10 classification in separate branches for two target classes, measuring the trace of the loss Hessian with respect to the output-layer parameters on fixed images of each target class. Across ten new initialisations, after 60 updates to the output layer with frozen features, the trace changed by $-0.152\%$ on average in branches where the target class share rose from $10\%$ to $19\%$, and by $+0.071\%$ in branches where it fell to $1\%$, relative to each branch's initial trace. These changes do not isolate the effect of class share from that of the assigned training images. We developed a forecast based on initial class gradients after examining the first experiment's results and then tested it prospectively without modification. For direct increases from $10\%$ to $19\%$, rather than increases after a deficit, its mean absolute error was $0.022$ percentage points on the initial-trace scale, compared with $0.152$ for a zero-change forecast and $0.104$ for equally weighted gradients. Using the assigned class proportions improved the forecast of how much the trace would change.
\end{abstract}

\section{Introduction}

When a class becomes more frequent, mean loss can change even for an unchanged classifier: only the class weights in the average have changed. A separate question is whether fine-tuning has changed the model itself. Has the mean quadratic contribution of small parameter perturbations to the loss on the original images of that class increased or decreased? To distinguish this response from simple reweighting, we measure the Hessian trace on the same images before and after fine-tuning.

The trace can distinguish model states that are similar in other respects. Layerwise loss-Hessian traces have been used to monitor the memorisation of noisy labels, using Hutchinson's stochastic estimator with Hessian--vector products \citep{bolshim_stochastic}. In an experiment involving a temporary learning-rate increase and gradient clipping, layerwise trace profiles differed at comparable losses even after the learning rate was restored; some gradient-based indicators were also informative \citep{bolshim_clipping_supplement}. The choice of operator matters, however: the spectrum and rank of the Hessian of a layer's response at fixed inputs \citep{bolshim_local} describe a different scalar function, not the loss. We therefore define explicitly the quantity used to study the class-conditional response to the composition of fine-tuning data.

We measure the exact Hessian trace of mean cross-entropy with respect to all weights and biases of the linear output layer. Separating representation learning from classifier learning \citep{kang} allows us to freeze the features. The trace is computed analytically from feature norms and current probabilities \citep{li}. It measures total curvature: for small, centred, isotropic perturbations of a given scale, it determines the mean quadratic contribution to the loss change. It is neither the full sensitivity, which also includes the linear gradient term, nor the worst-direction response, nor classification accuracy.

The relationship between the geometry of imbalance and optimisation \citep{rangwani} raises a further question: what can be known about this change before the first update? Probability adjustment using class priors \citep{saerens}, associated with the label-shift problem \citep{lipton}, describes a distributional change but not parameter movement over a short window. In the first experiment, using this adjustment as a forecast substantially overestimated the response. Examining that discrepancy led to a forecast based on initial class gradients. It was developed after the outcomes of the first experiment were known and then tested unchanged on ten new initialisations.

\section{Materials and Methods}

\subsection{Measured quantity}

The ten classes and corresponding outputs are indexed from zero to nine throughout. Let $h_i\in\mathbb{R}^{512}$ be the features of image $i$, let $x_i=(h_i^{\mathsf T},1)^{\mathsf T}\in\mathbb{R}^{513}$ be the augmented feature vector, and let $V=[W\mid b]\in\mathbb{R}^{10\times513}$ contain the output-layer weights and biases. Write $p_i=\operatorname{softmax}(Vx_i)$ and let $y_i$ be the true class label. Let $M$ index the measurement panel and $I_c\subset M$ its images of class $c$. For the single-image cross-entropy $\ell_i=-\log p_{i,y_i}$, the mean class-conditional trace is
\begin{equation}
\begin{aligned}
T_c
&=\operatorname{tr}\left[\nabla_{\operatorname{vec}V}^{2}
\frac{1}{|I_c|}\sum_{i\in I_c}\ell_i\right]\\
&=\frac{1}{|I_c|}\sum_{i\in I_c}\|x_i\|^{2}
\left(1-\sum_{j=0}^{9}p_{ij}^{2}\right),
\qquad I_c=\{i\in M:y_i=c\}.
\end{aligned}
\label{eq:trace}
\end{equation}
Equation~\eqref{eq:trace} averages traces over fixed images $I_c$ with unchanged weights $1/|I_c|$. Changing the training batches does not change the measurement images or their averaging weights. With the model held fixed, reweighting the classes could change the aggregate loss and its Hessian trace, but not $T_c$. With frozen features, a change in $T_c$ can be attributed to the change in probabilities after updating the output layer.

The trace is taken over the weights and biases of all ten outputs, not only the row for class $c$. The unit coordinate accounts for the bias, so that $\|x_i\|^2=1+\|h_i\|^2$. The second factor in Eq.~\eqref{eq:trace} describes probability uncertainty. For centred isotropic perturbations, the mean quadratic contribution to the loss equals the trace multiplied by half the variance of each perturbation coordinate; comparisons use the same small perturbation scale. Equation~\eqref{eq:trace} is analytically exact and requires neither the full-network Hessian nor stochastic probes. Logits for the current and final states on $M$ were obtained in FP32 in blocks of 100 images. Features and logits were converted to FP64 to compute probabilities, traces and subsequent comparisons.

On this fixed panel, we define the relative response, which is also the forecasting target, as
\begin{equation}
d_c=100\frac{T_{c,\mathrm{end}}-T_{c,\mathrm{start}}}
{T_{c,\mathrm{start}}}.
\label{eq:response}
\end{equation}
In prospective tests, the forecast $\widehat d_c$ is fixed before the first update. Equation~\eqref{eq:response} expresses the change as a percentage of the initial trace. The absolute error $|\widehat d_c-d_c|$ is measured in percentage points of this scale, not percentage points of classification accuracy. Each window uses its own initial trace as the denominator. Valid measurements require finite values and $T_{c,\mathrm{start}}>10^{-12}$.

\subsection{Static adjustment}

Let $q_{\mathrm{old},j}$ and $q_{\mathrm{new},j}$ denote the nominal training proportions of class $j$ before the composition change and in the assigned window, respectively. These are not estimates of the model's effective class priors. Method $\mathrm{F}$ replaces the current probabilities with adjusted probabilities and substitutes them into Eq.~\eqref{eq:trace}, keeping the features fixed:
\begin{equation}
\widetilde p_{ij}=
\frac{p_{ij}\,q_{\mathrm{new},j}/q_{\mathrm{old},j}}
{\sum_{k=0}^{9}p_{ik}\,q_{\mathrm{new},k}/q_{\mathrm{old},k}}.
\label{eq:prior}
\end{equation}
The Bayesian justification for Eq.~\eqref{eq:prior} assumes unchanged class-conditional input distributions and suitable initial posterior probabilities for the old class proportions. All proportions used here are positive. Nominal training proportions need not equal the model's effective prior distribution, and confidence does not guarantee calibration. The forecasting hypothesis tested here is that the adjusted probabilities approximate those of the model after 60 updates. The adjustment itself includes neither a learning rate nor a horizon; its Bayesian interpretation does not establish that forecasting hypothesis.

The zero-change forecast $\mathrm{B0}$ retains the initial trace.

\subsection{Forecasting the short-window response}

Let $e_c$ be the one-hot label vector for class $c$. The residual $p_i-e_{y_i}$ in the cross-entropy gradient is small for confident, correct predictions. Signed contributions to a given parameter can also cancel across images. The output-layer update therefore depends on the current probabilities and features, not just the target class's frequency. Unlike the uniform logit adjustment in $\mathrm{F}$, increasing a class's frequency does not in general imply adding the same increment to its logit across all measurement images. Forecast $\mathrm{G}$ uses this gradient information from the training data.

Let $J_c$ index the images of true class $c$ in a gradient bank drawn from the training set. The bank is used to estimate the update, not to measure the trace. At the start of the window, we compute the mean class gradient over all output-layer weights and biases:
\begin{equation}
g_c=\frac{1}{|J_c|}\sum_{i\in J_c}(p_i-e_c)x_i^{\mathsf T}.
\label{eq:gradient}
\end{equation}
The gradient in Eq.~\eqref{eq:gradient} has dimensions $10\times513$, like $V$. We make a single extrapolation from the initial state:
\begin{equation}
\widehat V=V-\eta K\sum_{c=0}^{9}q_cg_c,
\qquad
\widehat p_i=\operatorname{softmax}(\widehat Vx_i).
\label{eq:forecast}
\end{equation}
In Eq.~\eqref{eq:forecast}, $\eta=0.01$, $K=60$, and $q_c$ is the assigned proportion of class $c$ in the upcoming window. This is one extrapolation using the initial gradient, not another training run with 60 gradient updates. It does not use the future batches. We compute the trace using $\widehat p_i$ on the current features of $M$, then obtain the forecast from Eq.~\eqref{eq:response}. The forecasting control $\mathrm{G}_{\mathrm{bal}}$ uses the same features, initial class gradients and horizon, but assigns $q_c=0.1$ to every class. Comparing these forecasts tests the contribution of assigned rather than equal class weights applied to the same starting gradients, not a difference between trained branches.

The bank contains 5000 training images, 500 per class. Within each class, we selected the first 500 IDs in ascending order of the SHA-256 hash of the string \texttt{gradient04-v1:<ID>}, where \texttt{<ID>} is the decimal global ID; hash ties are resolved by ID. Selection is independent of the model and future batches. The same bank is used for development and the new test. It is disjoint from the measurement sets but may overlap with images used in subsequent training batches.

\subsection{Data and training}

We used CIFAR-10 \citep{krizhevsky} and an adapted ResNet-18 \citep{he}, with a $3\times3$ input convolution of stride 1, no initial max pooling, and a linear output layer with biases. Pixels were normalised as $u/127.5-1$. No random augmentation was used during either base training or fine-tuning. The task always involved all ten classes. Bird and ship were the two targets for changing class proportions and measuring traces in separate branches, not the only classes used for training.

Global IDs from 0 to 49\,999 refer to the sequential concatenation of the five original CIFAR-10 training files in numerical order. For classes 0 through 9 in sequence, the sorted IDs were permuted using a single NumPy \texttt{default\_rng} generator with seed 20261124. The first 4500 IDs per class were assigned to training, the next 250 to $M$, and the last 250 to a held-out set $H$. This produced disjoint sets of 45\,000, 2500 and 2500 images. The official test set was not used for method selection.

Each base model was trained for 20 balanced epochs, each consisting of 450 batches of size 100 with ten images per class. Each training ID appeared once per epoch. The initial learning rate was 0.1 and was reduced tenfold before epochs 13 and 18. Momentum was 0.9 and weight decay was 0.0005. Acceptance of the full set of ten base models required a median balanced accuracy on $M$ of at least 60\%; unsuccessful initialisations were not excluded. In the fine-tuning windows studied here, only $W$ and $b$ were updated, while features and running normalisation statistics remained fixed. A fresh stochastic gradient descent optimiser used a learning rate of 0.01, with no momentum or weight decay.

The primary window consisted of 60 updates with the target class proportion increased from 10\% to 19\%. Every batch of 100 contained exactly 19 target-class images and nine images from each other class. In the secondary deficit window, each batch contained exactly one target-class image and 11 images from each other class. These two direct branches started from the same weights of the corresponding base model. After the deficit window, another 60 steps with quotas of 19/9 were performed from that branch's own final state. Forecasts from this state used recomputed gradients of the current output layer. Features were reused after verifying that all other parameters and buffers were unchanged.

Within each class, IDs were randomly permuted and used without replacement within a window; image order within each batch was also shuffled. The direct increase and the increase after a deficit used the same batch array, with identical IDs and ordering. Differences between these branches therefore include the preceding 60 steps and different initial weights, rather than different training images or their order during the final window.

For $\mathrm{F}$, the increments to the target logit relative to the other logits in the three transitions are $\ln(19/9)$, $\ln(1/11)$ and $\ln(209/9)$. The last assumes adaptation to the deficit-window proportions, which short training need not achieve. Method $\mathrm{G}$ uses the assigned quotas for the upcoming window and gradients evaluated at that window's own initial state.

\subsection{Testing sequence and statistical analysis}

The first experiment included ten independent initialisations. Forecasts from $\mathrm{F}$ and $\mathrm{B0}$ were computed on $M$ and saved before the corresponding windows. After the experiment was completed and its analysis fixed, these same forecast values were compared, without recalculation, with the trace changes on $H$. Only then, with the first experiment's outcomes already known, was $\mathrm{G}$ developed and its forecasts computed retrospectively from the saved initial states. The held-out set $H$ was used neither to develop $\mathrm{G}$ nor in its prospective test; all subsequent results refer to $M$.

For the prospective test, we used the first ten prime numbers greater than 300 as random seeds. Base models were trained anew and training schedules were set, while the gradient bank, calculation method and coefficients were retained. Four methods were compared: $\mathrm{G}$, $\mathrm{G}_{\mathrm{bal}}$, $\mathrm{B0}$ and $\mathrm{F}$. Before the first update, their forecasts were fixed for 40 windows: 20 direct proportion increases and 20 deficits. After the deficits were completed, forecasts were computed for another 20 windows from their own final states, before any subsequent proportion-increase updates began. This gave 60 windows and 240 forecast values in total. No actual balanced branches were trained in the new test, so $\mathrm{G}_{\mathrm{bal}}$ remains a forecasting control.

Mean absolute error (MAE) was calculated by first averaging the two targets with equal weight within each initialisation, then averaging over the ten initialisations. Paired improvement was defined as the competitor's error minus the tested method's error, so a positive difference indicates an improvement. Confidence intervals used a paired percentile bootstrap with 10\,000 resamples of ten whole initialisations drawn with replacement, NumPy \texttt{default\_rng(47)}, and the 2.5th and 97.5th percentiles. Images, classes and windows were not treated as independent replicates. The intervals are conditional on the fixed dataset, two targets and gradient bank.

The original criterion required the lower confidence bound for the improvement of $\mathrm{F}$ over $\mathrm{B0}$ to exceed 1 percentage point in the direct-increase window on $M$. The new criterion, specified before training the new base models, addressed a different question: both lower bounds for the improvements of $\mathrm{G}$ over $\mathrm{B0}$ and over $\mathrm{G}_{\mathrm{bal}}$ had to be strictly positive. The same bootstrap resamples were used for these comparisons. If a model had been missing or a measurement invalid, the primary conclusion would not have been evaluated by reducing the set below ten. Intervals from retrospective development were descriptive.

\subsection{Numerical verification}

After the new test was completed, forecasts and measured changes were recomputed from the saved weights and features using a separate implementation, without importing the forecasting or evaluation modules of the main calculation. Mean class gradients were obtained by automatic differentiation of cross-entropy. The computational order and precision described above were retained for logits and traces. Paired intervals were recomputed using the same bootstrap scheme. No further training steps were performed.

\section{Results}

\subsection{Response on the original images}

All ten models were retained in the new test. Their median base-model balanced accuracy on $M$ was 87.36\%, against the 60\% threshold. In the direct branches from a common initial state, increasing the target proportion from 10\% to 19\% was accompanied by a mean trace change of $-0.15194\%$, while decreasing it to 1\% was accompanied by a change of $+0.07138\%$. These are changes on the same images relative to each branch's own initial state, not effects relative to balanced fine-tuning. The largest absolute response across all 60 windows was 0.25665\%. The proportion increase is the primary window and the deficit is secondary. To assess whether the magnitude of the shift could have been predicted, we first examine the static adjustment.

\subsection{Why the correct sign was not enough}

In the first experiment, median base-model balanced accuracy on $M$ was 87.22\%, and all ten models were retained. In the primary comparison, $\mathrm{F}$ predicted a mean trace decrease of 15.40\%, whereas the observed decrease was 0.124\%. The direction was correct in all 20 target--initialisation combinations, but the zero-change forecast was more accurate in every one. The mean paired improvement of $\mathrm{F}$ over $\mathrm{B0}$ was $-15.151$ percentage points, with a 95\% interval of $[-15.846,-14.346]$. This establishes worse performance, not merely failure to reach the threshold. Even a perfect forecast could have improved on zero by only 0.12422 percentage points on average, less than the original 1-percentage-point threshold. On $H$, the error difference was $-15.121$ percentage points, with an interval of $[-15.828,-14.308]$: a secondary assessment on different images from the same dataset.

The full adjustment increased the target-versus-rest log-odds by $\ln(19/9)=0.74721$ for every image. The actual mean increase on target-class images in $M$ was only 0.00718. To examine the scale of the discrepancy, after the experiment we considered the first assigned training batch for each target--initialisation combination, computing probabilities at the original weights before the first fine-tuning step. Across these 20 batches, the mean probability of the target class was 0.99668 on target-class images and 0.00051 on non-target images. On target-class images, the classifier was confident in the correct answer. The mean over all images was 0.18978, against a quota of 0.19. These conditional means describe the images in those training batches, not performance on new images.

With a constant quota and the optimiser used here, $\Delta b_c=\eta\sum_{t=1}^{60}(q_c-\overline p_{c,t})$, where $\overline p_{c,t}$ is the batch-mean probability before the update. The mean change in the target bias was approximately $2.03\times10^{-4}$; this yielded a reconstructed target-class probability of 0.18966, averaged over the batches in the window. A small mean signed residual does not imply that every gradient is small. We therefore examined the final displacement of the entire weight matrix $W$ separately: the mean relative Frobenius norm, $\|W_{\mathrm{end}}-W_{\mathrm{start}}\|_{\mathrm{F}}/\|W_{\mathrm{start}}\|_{\mathrm{F}}$, was 0.02634\%. This describes the weight matrix, not the full matrix $V=[W\mid b]$ including biases. The forecasting error concerns future probabilities, not the analytical trace or the Bayesian formula within its domain of applicability.

\subsection{Retrospective development and prospective testing}
\enlargethispage{\baselineskip}

The discrepancy suggested using initial gradients rather than a full static probability shift. In the retrospective calculation from the first experiment's initial states, $\mathrm{G}$ predicted a mean change of $-0.15131\%$, compared with the observed $-0.12422\%$. Its mean improvement over zero was 0.09122 percentage points, with a descriptive interval of $[0.07360,0.10978]$. After averaging over targets, improvements over both $\mathrm{B0}$ and $\mathrm{G}_{\mathrm{bal}}$ were positive in all ten initialisations. This motivated a new test; it did not replace one.

Table~\ref{tab:high} compares observed changes and forecast errors across the three stages.

\begin{table}[!htbp]
  \centering
  \caption{Observed change in the trace and forecast errors for a direct increase
    in the target share from 10\% to 19\%. The first two rows use the same
    observed outcomes. A dash indicates that the method was not evaluated at that stage.}
  \label{tab:high}
  \small
  \setlength{\tabcolsep}{5pt}
  \begin{tabularx}{\linewidth}{@{}X S[table-format=-1.5] S[table-format=2.5] *{3}{S[table-format=1.5]}@{}}
    \toprule
    & & \multicolumn{4}{c}{Mean absolute error (pp)} \\
    \cmidrule(l){3-6}
    Stage & {Observed (\%)} & {$\mathrm{F}$} & {$\mathrm{B0}$} & {$\mathrm{G}$} & {$\mathrm{G}_{\mathrm{bal}}$} \\
    \midrule
    Initial prospective test & -0.12422 & 15.27557 & 0.12422 & {\textemdash} & {\textemdash} \\
    Retrospective development on the same archive & -0.12422 & 15.27557 & 0.12422 & 0.03300 & 0.09176 \\
    Prospective test on new initialisations & -0.15194 & 14.69197 & 0.15194 & 0.02215 & 0.10395 \\
    \bottomrule
  \end{tabularx}
\end{table}

On the new initialisations, the unchanged $\mathrm{G}$ predicted a mean change of $-0.16956\%$, compared with the observed $-0.15194\%$ in the primary window. Its paired improvement over $\mathrm{B0}$ was 0.12979 percentage points, with a 95\% interval of $[0.10541,0.15269]$; the improvement over $\mathrm{G}_{\mathrm{bal}}$ was 0.08180 percentage points, with a 95\% interval of $[0.06708,0.09621]$. Both lower bounds were positive, satisfying the prespecified joint criterion. In each comparison, $\mathrm{G}$ was more accurate in all 20 target--initialisation combinations and, after averaging over targets, in all ten initialisations.

Relative to the zero-change forecast, error decreased by 85.4\%, but the absolute improvement was approximately 0.130 percentage points. A small overestimation of magnitude remained: the difference between the mean forecast and the mean observed change was $-0.01762$ percentage points. MAE is not the absolute value of this difference, because absolute values are taken before averaging.

The secondary windows are shown in Table~\ref{tab:secondary} and do not contribute to the primary decision. In both, $\mathrm{G}$ was more accurate than $\mathrm{B0}$ and $\mathrm{G}_{\mathrm{bal}}$ in all 20 target--initialisation combinations. Under the deficit, $\mathrm{G}_{\mathrm{bal}}$ predicted the opposite direction in every case: here, knowing the proportions changed not only the accuracy of the magnitude forecast but also its sign.

\begin{table}[!htbp]
  \centering
  \caption{Secondary windows in the prospective test on new initialisations.
    Changes and errors are relative to the initial trace of the corresponding window.}
  \label{tab:secondary}
  \small
  \setlength{\tabcolsep}{5pt}
  \begin{tabularx}{\linewidth}{@{}X *{2}{S[table-format=+1.5]} S[table-format=2.5] *{3}{S[table-format=1.5]}@{}}
    \toprule
    & \multicolumn{2}{c}{Trace change (\%)} & \multicolumn{4}{c}{Mean absolute error (pp)} \\
    \cmidrule(lr){2-3}\cmidrule(l){4-7}
    Target share & {Observed} & {$\mathrm{G}$ forecast} & {$\mathrm{F}$} & {$\mathrm{B0}$} & {$\mathrm{G}$} & {$\mathrm{G}_{\mathrm{bal}}$} \\
    \midrule
    $10 \to 1\%$ & +0.07138 & +0.07377 & 65.17277 & 0.07138 & 0.01155 & 0.11937 \\
    $1 \to 19\%$ after a deficit & -0.15318 & -0.17084 & 52.99415 & 0.15318 & 0.02219 & 0.10427 \\
    \bottomrule
  \end{tabularx}
\end{table}

Both $\mathrm{G}$ and $\mathrm{F}$ predicted the correct sign in all 60 windows of the new test. Again, the main difference concerned magnitude: in the direct-increase window, $\mathrm{F}$ predicted $-14.84391\%$ rather than $-0.15194\%$. The largest absolute forecast from $\mathrm{G}$ was 0.30308\%.

The separate implementation reproduced all 240 forecasts and 60 measured changes, as well as both primary confidence intervals. For forecasts and measured changes, the maximum absolute discrepancy from the main calculation was below $3\times10^{-13}$ percentage points. This verifies the numerical procedure for the completed experiment; it is neither a new training series nor additional independent replication.

\section{Discussion}

With frozen features, only the probability-uncertainty factor in Eq.~\eqref{eq:trace} changes. A decrease in trace means a smaller mean quadratic contribution of output-layer perturbations to the loss for the selected class; an increase means a larger contribution at the same perturbation scale. This is a geometric characteristic, not an assessment of classification performance.

A static distributional adjustment and parameter movement over 60 steps are different transformations. Even if the assumptions of Eq.~\eqref{eq:prior} hold, the new class proportions do not specify how quickly a trained model will approach the adjusted probabilities. In the first experiment, confident predictions on the images in the assigned training batches, a small mean signed residual and a separately measured small weight displacement were consistent with a much smaller response than $\mathrm{F}$ predicted, but do not provide an exhaustive causal explanation. Initial class gradients offer a testable approximation to parameter movement over this short window.

Method $\mathrm{F}$ uses the current probabilities and features on $M$ together with nominal proportions, while $\mathrm{B0}$ retains the initial trace. Method $\mathrm{G}$ additionally requires a labelled training bank, its features and the calculation of initial class gradients. Method $\mathrm{G}_{\mathrm{bal}}$ uses the same starting class gradients but replaces the assigned proportions with equal weights. The advantage of $\mathrm{G}$ over this control shows the forecasting value of assigned rather than equal class weights applied to those gradients. This compares two forecasts, not a forecast with an actually trained balanced branch.

The tested regime is narrow: 60 steps, output-layer updates only, known proportions and two target classes. The effect of proportions was not isolated from that of the particular training images and their order. A constant initial gradient does not capture trajectory curvature or the randomness of specific batches. The similarity between a direct increase and an increase after a deficit does not rule out a role for training history, since their initial weights differ.

The test on new initialisations used the same images, split, architecture and gradient bank. With ten initialisations, the bootstrap intervals are approximate and do not include uncertainty from selecting the bank or the panel. Saving forecasts before their corresponding updates establishes the order of calculations within the experiment but does not substitute for external timestamp registration.

Schedule selection, accuracy improvements, resource savings and operation with unknown proportions were not tested. A substantive next question is the horizon over which the initial extrapolation remains useful as the learning rate and response magnitude increase. Training the full network would also require accounting for feature movement. Changing the loss function may alter the problem itself.\footnote{For scalar regression with a linear output and loss $\tfrac12(\widehat y-y)^2$, the Hessian trace with respect to weights and bias is $1+\|h\|^2$. With fixed features, training only the output layer does not change it.}

\section{Conclusion}

Fine-tuning changed class-conditional curvature on the original images: increasing the target proportion was accompanied by a decrease in trace, whereas decreasing the proportion was accompanied by an increase in trace, relative to each branch's own initial state. This is a geometric response of the model, not a reweighting of images. It was observed for two CIFAR-10 target classes in ten-class classification with 60 output-layer-only training steps.

In the tested regime, the response magnitude in the primary direct-increase window could be estimated in advance. The full proportion adjustment predicted the correct sign but overestimated the magnitude. A forecast based on initial class gradients was developed after the outcomes of the first experiment were known, then tested unchanged on ten new initialisations. In the primary window, $\mathrm{G}$ was more accurate than both control forecasts: zero change and equally weighted gradients. Within this regime and window, known future proportions combined with the model's initial class gradients improved forecasting.

\section*{Funding}
This work was carried out under the state assignment of the Ministry of Science
and Higher Education of the Russian Federation (project FSER-2025-0004).

\bibliographystyle{unsrtnat}
\providecommand{\doi}[1]{\href{https://doi.org/#1}{doi:\,\nolinkurl{#1}}}
\bibliography{references}
\end{document}